\documentclass[10pt,twocolumn,letterpaper]{article}

\usepackage{iccv}
\usepackage{times}
\usepackage{epsfig}
\usepackage{graphicx}
\usepackage{amsmath}
\usepackage{amssymb}

\usepackage{enumitem}
\usepackage{IEEEtrantools}
\usepackage{mathtools}
\usepackage{url}
\usepackage{caption}
\usepackage{subcaption}
\usepackage{multirow}
\usepackage{tabularx}
\usepackage{booktabs}
\usepackage{nccmath}

\DeclarePairedDelimiter\norm{\lVert}{\rVert}%

\usepackage[pagebackref=true,breaklinks=true,letterpaper=true,colorlinks,bookmarks=false]{hyperref}

\iccvfinalcopy 

\def\iccvPaperID{7223} 

\ificcvfinal\fi

\begin{document}

\title{Real-time Image Enhancer via Learnable Spatial-aware 3D Lookup Tables}

\author{Tao Wang\thanks{Authors contributed equally}, Yong li\footnotemark[1], Jingyang Peng\footnotemark[1], Yipeng Ma, Xian Wang, Fenglong Song\thanks{Corresponding author}, Youliang Yan\footnotemark[2]\\
Huawei Noah's Ark Lab\\
{\tt\small {\{wangtao10,liyong156,pengjingyang1,mayipeng,wangxian10,songfenglong,yanyouliang\}}@huawei.com}

%
}

\maketitle
\thispagestyle{empty} 
\ificcvfinal\pagestyle{empty}\fi

\begin{abstract}
   Recently, deep learning-based image enhancement algorithms achieved state-of-the-art (SOTA) performance on several publicly available datasets. However, most existing methods fail to meet practical requirements either for visual perception or for computation efficiency, especially for high-resolution images. In this paper, we propose a novel real-time image enhancer via learnable spatial-aware 3-dimentional lookup tables(3D LUTs), which well considers global  scenario and local spatial information. Specifically, we introduce a light weight two-head weight predictor that has two outputs. One is a 1D weight vector used for image-level scenario adaptation, the other is a 3D weight map aimed for pixel-wise category fusion. We learn the spatial-aware 3D LUTs and fuse them according to the aforementioned weights in an end-to-end manner. The fused LUT is then used to transform the source image into the target tone in an efficient way. Extensive results show that our model outperforms SOTA image enhancement methods on public datasets both subjectively and objectively, and that our model only takes about 4ms to process a 4K resolution image on one NVIDIA V100 GPU.

\end{abstract}

\begin{figure}[tb]
	\centering
	\includegraphics[width=0.9\linewidth]{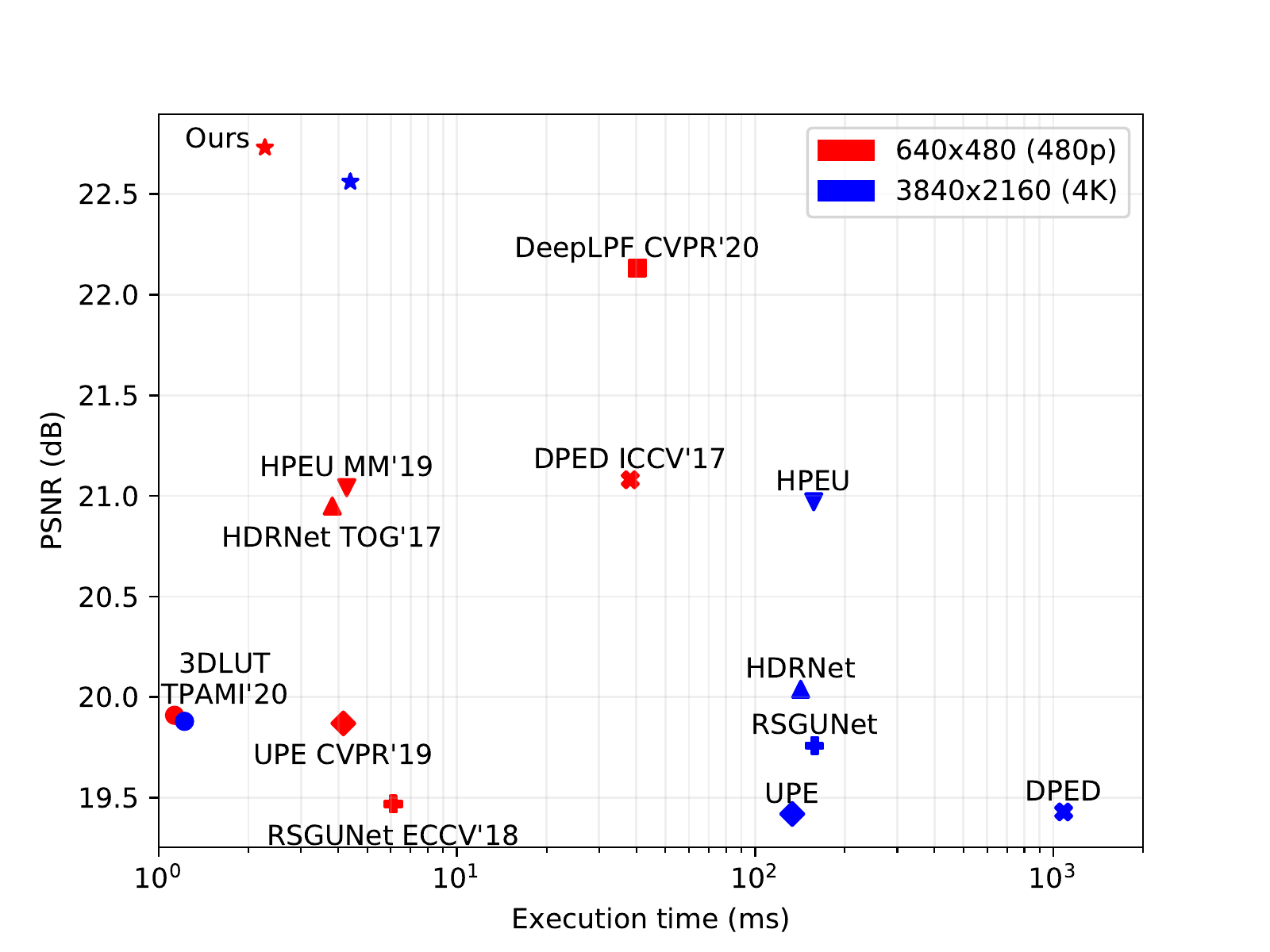}
	\caption{Performance and efficiency on HDR+ burst photography dataset of different methods for 480p ($640\times 480$) and 4K ($3840\times2160$) resolution on NVIDIA V100 GPU. Our method achieves the highest PSNR and the second fastest execution speed. DeepLPF~\cite{moran2020deeplpf} is out of memory on 4K resolution. \label{fig_psnr_time_scatter}}
\end{figure}

\begin{figure*}[!h]
	\centering
	\includegraphics[width=0.95\textwidth]{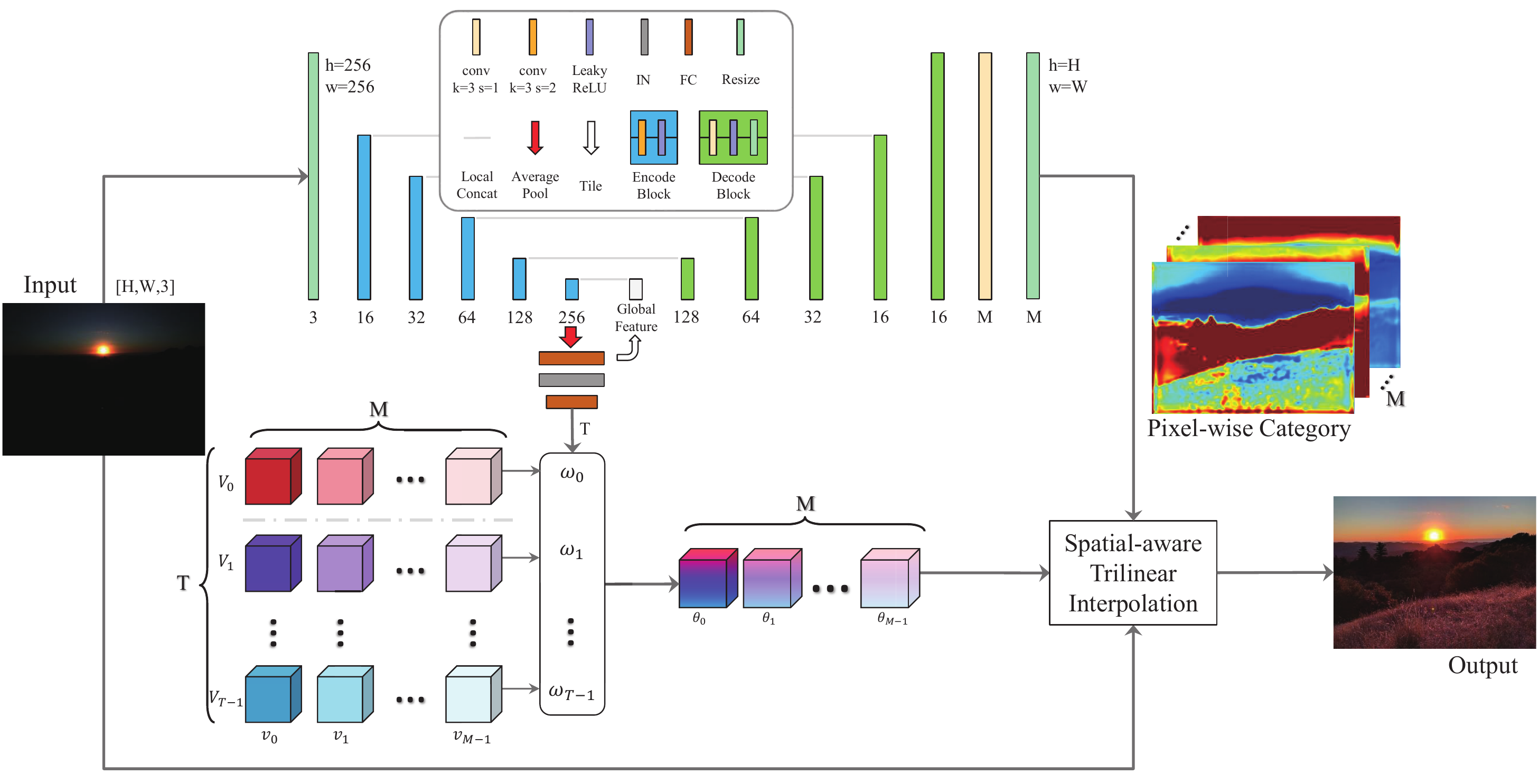}
	\caption{Overview of our proposed framework. It consists of multiple spatial-aware 3D LUTs (i.e., $T$ spatial-aware 3D LUTs, each with $M$ basic 3D LUTs selected by $M$-channel pixel-wise category information.), a self-adaptive two-head weight predictor, and interpolation for spatial-aware 3D LUTs. The weight predictor takes down-sampled images as input and generates two outputs. One is a 1D weight vector used for image-level scenario adaptation, the other is a 3D weight map aimed for pixel-wise category fusion, enabling our LUT-based enhancer with image-adaptive spatial-aware ability.\label{fig_architecture}}
\end{figure*}

\section{Introduction}

Recently, many deep learning-based approaches have been proposed and achieved SOTA results~\cite{huang2018range,moran2019curl,zeng2020learning,dong2015image,vu2018fast,mei2020pyramid,zhang2017beyond,tai2017memnet,zhang2020residual} in the field of computational imaging. However, complex network architecture and high computation overheads prevent them from real-time processing. Figure~\ref{fig_psnr_time_scatter} shows the comparison of performance and efficiency (i.e., execution time) of several network architectures on HDR+ Burst Photography dataset~\cite{hasinoff2016burst}. Most existing methods cannot produce visually pleasant results in real time.

Considering both performance and efficiency, it is still a big challenge for image enhancement due to the diversity of capture scenarios. Recently, many hybrid methods~\cite{huang2019hybrid,zeng2020learning,zheng2020image}, which combine image prior in traditional approaches and multi-level features in deep learning-based approaches, are proposed and achieve SOTA performance.  ~\cite{zeng2020learning} proposes a new image enhancement method with good image quality, high computation efficiency and low memory consumption. However, as the limitations pointed out by authors, it works simply based on pixel values, without considering local information. This may produce less satisfactory results in local areas. For example, as shown in Figure~\ref{fig_sota_mit5k}, local contrast is limited in some results captured in high dynamic range scenes. In addition, there are also some color distortion and artifacts as shown in Figure~\ref{fig_sota_hdrplus}.

To solve these issues, we present a novel CNN-based image enhancement approach, where spatial information is introduced to traditional 3D lookup tables to boost its performance.  Particularly, T spatial-aware 3D LUTs, each of which is a set of M basic learnable 3D LUTs, and a two-head weight predictor are trained simultaneously under a new loss function to balance well between details, colors, and visual perception. The weight predictor has two outputs. One is a 1D weight vector with global information used for integration of different LUTs on dimension T, which is called image level scenario adaptation. The other is a 3D weight map with pixel-wise category information aimed for combination of multiple LUTs on dimension M,  which is named pixel-wise category fusion.   Enhanced images are obtained by fusion of spatial-aware 3D LUTs according to the aforementioned two kinds of weights.  In addition, our approach only takes about 4 ms to process an image of 4K resolution on  NVIDIA V100 GPU platform.

The main contributions are summarized as follows:
\begin{itemize}[nosep]
	\item We propose a spatial-aware 3D LUTs architecture by constructing multiple basic 3D LUTs and introducing two-head weight predictor. This architecture makes it more robust in local enhancement. 
	\item We design a two-head weight predictor which  learns image-level scenario and pixel-wise category information with low computation overheads. Such weight information combined with spatial-aware 3D LUTs effectively improves the performance of image enhancement, and balances well between detail, color and perception under the supervision of our loss functions.
	\item We conduct extensive experiments to compare our approach with existing methods on two public datasets. Results demonstrate advantages of our approach quantitatively and qualitatively on performance and efficiency.
\end{itemize}

\section{Related Work}
Existing learning-based approaches can be broadly divided into three categories, namely pixel-level, patch-level, and image-level methods.

\textbf{Pixel-level methods.} This kinds of methods adopt CNN to extract features from input images of initial size and reconstruct every pixel from dense pixel-to-pixel mapping or transformation operations. This kind of approaches have made great breakthroughs and achieved SOTA performance in many image enhancment tasks
~\cite{jiang2019enlightengan,Chen2018Retinex,zhang2019kindling,chen2018deep,zamir2020learning,moran2020deeplpf,chen2018learning}.%
~\cite{ignatov2017dslr} proposes a residual CNN architecture as enhancer to learn the pixel-wise translation function between low-quality cellphone images and high-quality 
Digital Single-Lens Reflex (DSLR) images.~\cite{chen2018deep,jiang2019enlightengan,chen2018learning,huang2019hybrid} all employ UNet-style structure originated from~\cite{ronneberger2015u} for different image quality enhancement tasks. Despite their SOTA performance, these dense pixel-wise feature extraction and regeneration methods are too heavy to be used for practical applications, especially for high resolution input images~\cite{zeng2020learning}.

\textbf{Patch-level methods.} These methods generate compressed features from a down-sampled input image. Different parts of features are then applied on different local input patches to reconstruct the enhanced image.~\cite{gharbi2017deep} extracts local and global features as a bilateral grid in low resolution, and then applies interpolation according to the grid and a learned feature map of full resolution. Based on the same interpolation operation,~\cite{wang2019underexposed} learns a full resolution illumination map to retouch the input image. Wu et al.~\cite{wu2018fast} introduce the guided filter proposed in~\cite{he2010guided}, and they build a trainable guided filtering layer and plug it in the network for up-sampling the enhanced low resolution image. Although patch-level methods perform well both on computation and memory consumption, they still overload hardware resources, especially for ultra-high resolution images.

\textbf{Image-level methods.} These approaches have the highest computation efficiency and lowest memory consumption. They calculate global scaling factors or mapping curves from a down-sampled input image, which are then applied on the whole input image for enhancement. ~\cite{zeng2020learning} propose image-adaptive 3D LUTs for efficient image enhancement and it takes only 1.66 ms to process a 4K image on NVIDIA Titan RTX GPU. However, it is hard to ensure robustness since spatial information is not considered, which may easily result in low local contrast or even wrong color in some local areas, as shown in Figure~\ref{fig_sota_mit5k} and Figure~\ref{fig_sota_hdrplus}.

\section{Methodology}

In this section, we present our network framework and loss functions in detail. Figure~\ref{fig_architecture} illustrates fundamental modules of our network architecture, consisting of multiple spatial-aware 3D LUTs, a self-adaptive two-head weight predictor and spatial-aware trilinear interpolation.

\subsection{Network architecture}

\textbf{Spatial-aware 3D LUTs}. 3D LUT is an effective color mapping operator, which contains two basic operations: lookup and interpolation. For simplicity of description, we do not describe the interpolation operation in 3D LUT, but simplify it to lookup only in this subsection.  

Equation~\ref{eq_lut_mapping} indicates the mapping function. In RGB color domain, a classical 3D LUT is defined as a 3D cube which contains $N^3$ elements, where $N$ is the number of bins in each color channel. Each element defines a pixel-to-pixel mapping $\mu^c{(i,j,k)}$, where  $i,j,k=0,1,\dots,N-1$, abbreviated as $i,j,k\in\mathbb{I}^{N-1}_0$ in the following section, are elements' coordinates within 3D LUT and $c$ indicates one of channels. Inputs of the mapping are RGB colors $\{I^r_{(i,j,k)}, I^g_{(i,j,k)}, I^b_{(i,j,k)}\}$, where $i,j,k$ are indexed by the corresponding RGB value, and output is the pixel value after mapping for channel $c$, as in Equation~\ref{eq_lut_mapping}. $O^c$ is output for 3D LUT with $c\in\{r, g, b\}$, and $r,g,b$ is the color value for red, green, blue channel respectively.
\begin{eqnarray}
	O^c_{(i,j,k)} = \mu^c(I^r_{(i,j,k)},I^g_{(i,j,k)},I^b_{(i,j,k)}) \IEEEyesnumber
	\label{eq_lut_mapping}
\end{eqnarray}

\begin{figure}[tb]
	\includegraphics[width=\linewidth]{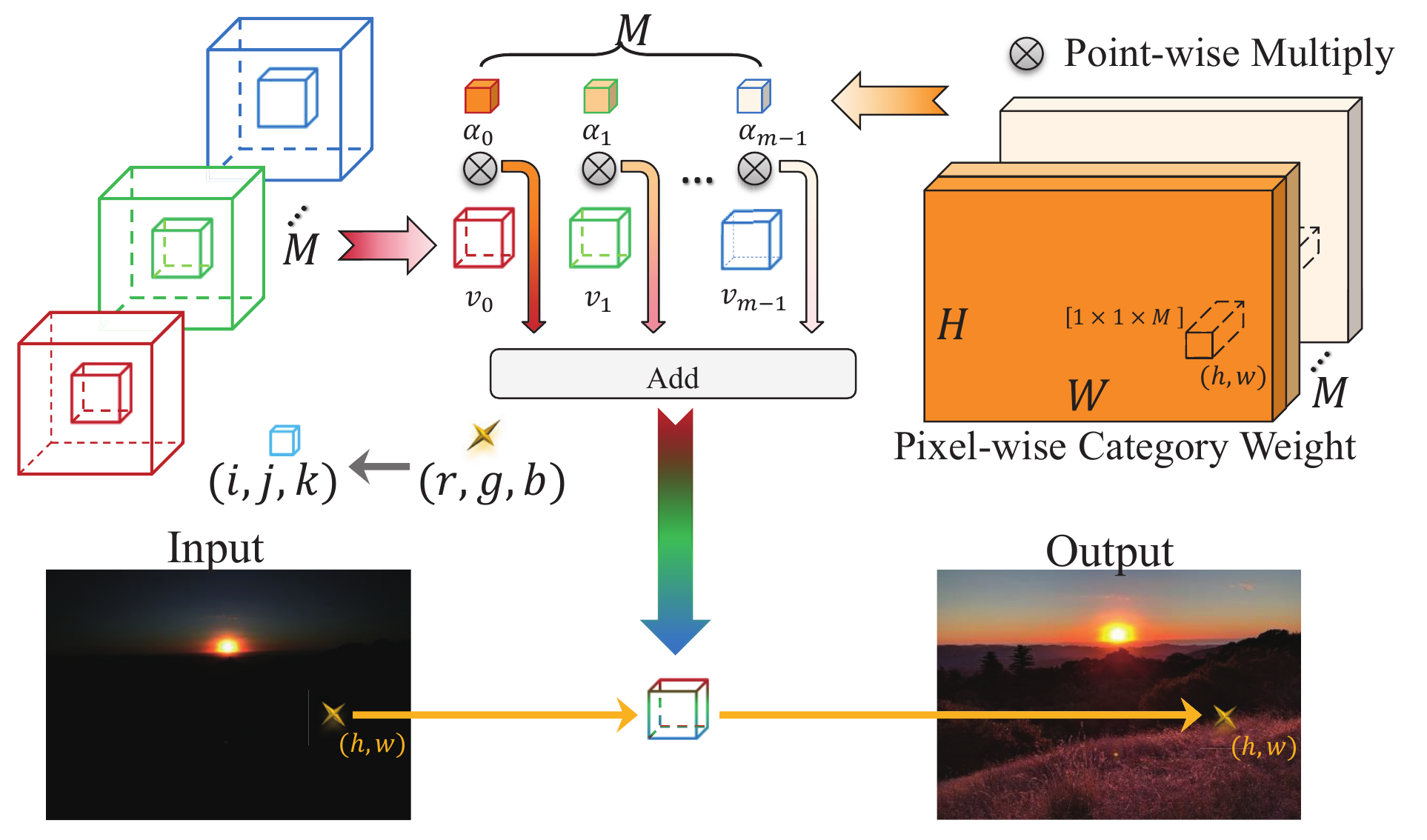}
	\caption{Visualization of our spatial-aware 3D LUTs. An input pixel at location ($h,w$) with pixel value ($r,g,b$) corresponds to $M$ LUT cells $\{v_t\}$ and a pixel-wise weight map with size $1\times 1 \times M$, where the 3DLUT key ($i,j,k$) is indexed from ($r,g,b$) value. The final fused LUT cell, generated by the weighted sum result of M basic LUTs, is used to obtain the output.\label{fig_3d_lut}}
\end{figure}

Obviously, mapping for traditional 3D LUT depends merely on pixel values, but fails to consider spatial information. In other words, the transformation is only sensitive to pixel values, and discards pixels’ spatial information. We propose new spatial-aware 3D LUTs involving $M$ traditional 3D LUTs, each of which represents a kind of mapping. For the final result, our method adaptively fuses multiple LUTs according to  pixel-wise weight map. As shown in Equation~\ref{eq_spa_lut_mapping},  $\phi^{h,w,c}$ is the entire mapping, $\nu^c$ is a mapping for the $m$-th LUT and $\alpha^{h,w}=\{\alpha^{h,w}_{m}|h\in\mathbb{I}^{H-1}_0, w\in\mathbb{I}^{W-1}_0, m\in\mathbb{I}^{M-1}_0\}$ is a spatial-aware pixel-wise weight map for $M$ 3D LUTs at location $(h,w)$. 

\begin{align}
	&O^{h,w,c}_{(i,j,k)} = \phi^{h,w,c}(I^r_{(i,j,k)},I^g_{(i,j,k)},I^b_{(i,j,k)}, \alpha^{h,w}) \nonumber\\
	&=\sum_{m=0}^{M-1}\alpha^{h,w}_{m}\nu^c(I^r_{(i,j,k)},I^g_{(i,j,k)},I^b_{(i,j,k)},m) \nonumber \\
	&=\sum_{m=0}^{M-1}\alpha^{h,w}_{m}O^{m,c}_{(i,j,k)} 
	\label{eq_spa_lut_mapping}
\end{align}
where $O^{h,w,c}_{(i,j,k)}$ is the final spatial-aware result and $O^{m,c}_{(i,j,k)}$ is the mapping result of the $m$-th 3D LUT.

Note that pixels are adaptively classified into different categories through an end-to-end learning approach according to color, illumination, semantic and other information. This generalizes our model to different use cases and promotes its learning ability. Figure~\ref{fig_3d_lut} visualizes our spatial-aware 3D LUTs.

We use $V=\{\phi^{h,w,c}{(i,j,k,\alpha^{h,w})}\}$ to represent a set of all mappings in spatial-aware 3D LUTs. Thus, $\mathrm{Y}=V(\mathrm{X}, \mathrm{A})$ indicates applying spatial-aware 3D LUTs on input image $\mathrm{X}$. $\mathrm{A}=\{\alpha^{h,w}_{m}|h\in\mathbb{I}^{H-1}_0,w\in\mathbb{I}^{W-1}_0, m\in\mathbb{I}^{M-1}_0\}$ is the pixel-wise category information, which is introduced in next part.

\textbf{Self-adaptive two-head weight predictor.} We propose a self-adaptive two-head weight predictor to support image-adaptive spatial-aware 3D LUTs. Upper part of Figure~\ref{fig_architecture} shows its framework, which is a UNet-style backbone with two outputs. The first one is a 1D weight vector with $T$ probabilities $\{\omega_t|t\in\mathbb{I}^{T-1}_0\}$, where $T$ is scene number. These $T$ probabilities are used for scene adaptation. We assume that the scene is a global feature, and its probability can be expressed by a single value in the probability vector. With these probabilities, a scenario-adaptive 3D LUT can be jointly leaned by $T$ spatial-aware 3D LUTs. For an input image $\mathrm{X}$, the final enhancement result $\mathrm{Y}$ can be expressed as follows. In the following experiments, we set $T=3$ according~\cite{zeng2020learning}.
\begin{eqnarray}
	\mathrm{Y} = \sum_{t=0}^{T-1} \omega_t*V_t(\mathrm{X}, \mathrm{A})
	\label{eq_img_merge}
\end{eqnarray}

The second output is an $M$-channel 3D weight map with $H \times W \times M$ probabilities $\mathrm{A}=\{\alpha^{h,w}_{m}|h\in\mathbb{I}^{H-1}_0,w\in\mathbb{I}^{W-1}_0, m\in\mathbb{I}^{M-1}_0\}$, as shown in Figure~\ref{fig_architecture}. Each channel corresponds to fusion weight for specific LUTs as shown in Figure~\ref{fig_3d_lut}. With the pixel-wise weight information, spatial feature is fused to 3D LUTs, which greatly promotes enhancement result in many aspects, e.g., local contrast and saturation. Detailed results are analyzed in Section~\ref{sec_experiments}. 

Our weight predictor takes resized low resolution images as inputs, enabling it to process arbitrary size images in real time. Moreover, the Encoder-Decoder architecture increases the receptive field size, which is powerful in generating pixel-wise category feature.

\textbf{Spatial-aware trilinear interpolation.} Considering the efficiency and performance, trilinear based interpolation is used in our method to improve the smoothness of the enhanced result.For detailed derivation, please refer to the supplementary material. Owning to the spatial-aware attribute of the pixel-wise category weight map $\alpha_m^{h,w}$, the interpolation is defined as spatial-aware trilinear interpolation.

\subsection{Loss function}
Our loss function consists of MSE Loss, Smooth Loss ~\cite{zeng2020learning}, Monotonicity Regularization Loss ~\cite{zeng2020learning}, Color Difference Loss and Perception Loss. MSE Loss  ($L_{r}$) ensures content consistency of generated image. Smooth Loss ($L_{s}$) and Monotonicity Regularization Loss ($L_{m}$) are introduced to ensure LUTs' smoothness and reduce artifacts.

Additionally, in order to promote enhancement quantitatively and perceptually, we introduce Color Difference Loss ($L_c$) and Perceptual Loss ($L_p$) to the optimization process.

\textbf{Color Difference Loss.} To measue the color distance and encourage the color in the enhanced image to match that in the corresponding learning target,  we use CIE94 in LAB space as our color loss. Detailed description can be found in~\cite{deltae} and supplementary material.

\begin{align}
	L_{c}&=\sqrt{\Delta L^2 + \left( \frac{\Delta C}{S_{C}} \right)^2 + \left( \frac{\Delta H}{S_{H}} \right)^2 + \epsilon} \IEEEyesnumber
	\label{eq_color_loss}
\end{align}

\textbf{Perception loss.}  LPIPS loss~\cite{zhang2018unreasonable} is chosen to improve the perceptual quality of the enhanced image.
\begin{equation}
	L_p=\sum_l \frac{1}{H^{l}W^{l}}\sum_{h=1,w=1}^{H^{l},W^{l}}\norm*{ \hat{y}^l_{hw} - y^l_{hw} }^2_2
\end{equation}
where $l$ is the layer chosen to calculate lpips loss, and $\hat{y}^l, y^l$ is the corresponding ground truth features and enhanced features on a pre-trained AlexNet.

Finally, the loss function is defined as a weighted sum of different losses with following coefficients.
\begin{equation}
	L = L_{r} + 0.0001*L_{s} + 10*L_{m} + 0.005*L_{c} + 0.05*L_p
\end{equation}

\begin{table*}[!t]
	\centering
	\begin{tabularx}{\textwidth}{>{\hsize=.30\hsize\linewidth=\hsize}X 
			>{\hsize=.05\hsize\linewidth=\hsize \centering\arraybackslash}X
			>{\hsize=.05\hsize\linewidth=\hsize \centering\arraybackslash}X
			>{\hsize=.24\hsize\linewidth=\hsize \centering\arraybackslash}X
			>{\hsize=.24\hsize\linewidth=\hsize \centering\arraybackslash}X
			>{\hsize=.1\hsize\linewidth=\hsize \centering\arraybackslash}X
			>{\hsize=.09\hsize\linewidth=\hsize \centering\arraybackslash}X
			>{\hsize=.08\hsize\linewidth=\hsize \centering\arraybackslash}X
			>{\hsize=.08\hsize\linewidth=\hsize \centering\arraybackslash}X
			>{\hsize=.08\hsize\linewidth=\hsize \centering\arraybackslash}X
		}
		
		\toprule[1pt]
		\multirow{2}{100em}{Method(T,M)} & \multicolumn{4}{c}{Configuration} & \multirow{2}{100em}{GFLOPS} & \multirow{2}{100em}{\#Params} & \multirow{2}{100em}{PSNR $\uparrow$}  & \multirow{2}{100em}{SSIM $\uparrow$} & \multirow{2}{100em}{LPIPS $\downarrow$} \\
		\cmidrule(lr){2-5} 
		& CNN & input & Weight predictor & \#3DLUTs &   \\ 
		
		\hline 
		3DLUT(3,0)  & ~\cite{zeng2020learning} & 480p & 1D & 3$\times$ basic   & 0.206 & 539K  & 19.91 & {0.6567} & {0.2455} \\
		3DLUT(30,0) & ~\cite{zeng2020learning} & 480p & 1D & 30$\times$ basic  & 0.209 & 3.72M & 20.29 & {0.6614} & {0.2306} \\
		Ours(30,0) & ours & 480p & 1D & 30$\times$ basic  & 0.228 & 3.74M & 20.38 & {0.6888} & {0.2249} \\
		Ours(0,30) & ours & 480p & 3D & 1$\times$ spatial-aware  & 1.934 & 4.48M & \textcolor{blue}{22.52} & \textcolor{blue}{0.7316} & \textcolor{blue}{0.1878} \\
		Ours(3,10) & ours & 480p & 1D\&3D & 3$\times$ spatial-aware  & 1.114 & 4.52M & \textcolor{red}{22.73} & \textcolor{red}{0.7420} & \textcolor{red}{0.1580} \\
		\hline
		Ours$^{\star}$(3,10) & ours & 4K & 1D\&3D & 3$\times$ spatial-aware  & 8.111 & 4.52M & 22.56 & {0.6996} & {0.2808} \\
		Ours-noresize$^{\star}$(3,10) & ours & 4K & 1D\&3D & 3$\times$ spatial-aware  & 113.792 & 4.52M & {22.65} & {0.7323} & {0.2142} \\		
		\bottomrule[1pt]
	\end{tabularx}
	\caption{Ablation study for different combinations of CNN weight predictor and 3DLUTs. A spatial-aware 3D LUTs is composed with M basic 3D LUTs. }
	\label{tab_num_classifiers}
\end{table*}

\section{Experiments}
\label{sec_experiments}

\textbf{Datasets.} We evaluate our method on two publicly available datasets: MIT-Adobe FiveK~\cite{bychkovsky2011learning} and HDR+ burst photography~\cite{hasinoff2016burst}. Since~\cite{zeng2020learning} achieved the SOTA performance on both dataset and also published its 480p dataset (only 480p, w/o full resolution), we directly take their released 480p dataset for performance evaluation.  We also construct two new dataset for further comparison. One is full resolution MIT-Adobe FiveK dataset. The ExpertC images are used as the groundtruth while the input DNG images  are automatically converted to PNG images as input. We use the same filelist as ~\cite{zeng2020learning} for training and testing. The other is 480p and full resolution HDR+ dataset. Our input images are merged DNG images (i.e., merge.dng) post-processed by python rawpy library with automatic white balance, while ground truth images are kept as the software output (i.e., final.jpg). Since most scenes are not well aligned in the original dataset, we conduct manual comparison and remove image pairs with large offset. In this way, we construct a dataset with 2041 image pairs.  Finally, we randomly split image pairs in the dataset into two subsets: 1837 image pairs for training and the rest 204 pairs for testing. Since the number of the released 480p HDR+ dataset by ~\cite{zeng2020learning} is relatively small(675 pairs), we also construct our 480p HDR+ dataset with the short side resized to 480 pixels and long side proportionally.

\textbf{Evaluation metrics.} We employ three common-used metrics (i.e., PSNR, SSIM and LPIPS) to quantitatively evaluate the performance of different methods. Generally speaking, a higher PSNR/SSIM and lower LPIPS means better results.

\textbf{Application settings.} We implement our network with pytorch~\cite{paszke2017automatic} and train all modules on a NVIDIA V100 GPU for 400 epochs with a mini-batch of 1. The entire network is optimized using standard Adam~\cite{kingma2014adam} with a cosine annealing learning rate with amplitude 2e-4 and period 20 epochs. The spatial-aware trilinear interpolation is accelerated via customized CUDA code.

\subsection{Ablation study}
To demonstrate the effectiveness of different components of our approach, we conduct several ablation studies on our HDR+ dataset.

\begin{figure}[tb]
	\centering
	\includegraphics[width=0.9\linewidth]{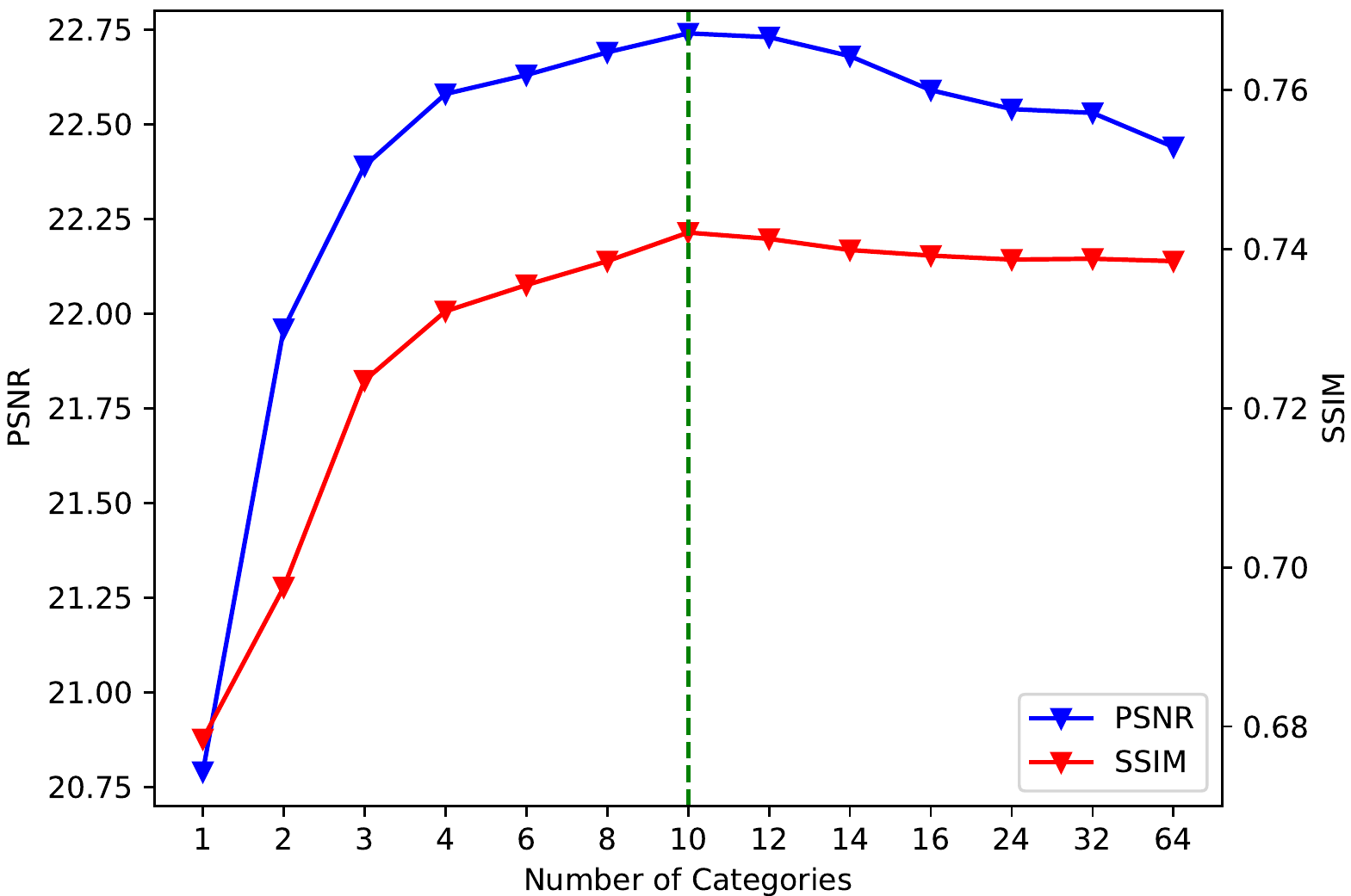}
	\caption{Ablation studies on the number of categories ($M$).\label{fig_num_of_refmaps}}
\end{figure}

\textbf{Number M of LUTs.} We assess the performance of different settings to determine the number of pixel-wise category for spatial-aware 3D LUTs with $T=3$. Figure~\ref{fig_num_of_refmaps} shows models' performance with different pixel-wise category number ($M$) from $M=1$ to $M = \{2, 3, 4, 6, 8, 10, 12, 14, 16, 24, 32, 64\}$. We can see an evident improvement by increasing $M$ from $1$ to $10$, but minor improvement or even deterioration if $M$ is further increased. Therefore, $M$ is set to $10$ in all our following experiments. 

\textbf{Two-head weight predictor.} To further demonstrate the contribution of our whole architecture, we continuously conduct the following experiments with different combination of CNN weight predictor and 3D LUTs. $(t,m)$ replesents the CNN configuration with  $T=t, M=m$.

As shown in Table~\ref{tab_num_classifiers}, directly increasing the number of LUTs based on Zeng’s ~\cite{zeng2020learning} method cannot improve performance effectively. Both our two-head weight predictor and spatial-aware 3D LUTs are important. Our 1D weight(i.e., ours(30,0)) cannot work well alone when it is used alone, even if the number of LUTs is the same as our final configuration. The 3D weight (i.e., ours(0,30)) shows effectiveness in performance improvement when cooperated with our spatial-aware interpolation. When both 1D weight and 3D weight are utilized (i.e.,ours(3,10)), our model performs better, which shows a total of 2.82 dB improvement in PSNR when compared with the original one. Additionally, our method can also work well on full resolution image(i.e.,ours$^{\star}$(3,10)), only with 0.17dB degradation in PSNR.  After deleting the first and last resize operation in CNN weight predictor(i.e.,ours-noresize$^{\ast}$(3,10)), it can achieve 0.09 dB improvement, but the computation FLOPS improves from 8.111G to 113.79G. Therefore, we use ours(3,10) with resize operation as shown in Figure~\ref{fig_architecture} as our final architecture.

\begin{table}[tb]
	\begin{center}
		\begin{tabular}{c c c c c c}
			\toprule[1pt]
			$L_b$ & $L_c$ & $L_p$ & PSNR $\uparrow$ & SSIM $\uparrow$ &LPIPS $\downarrow$ \\
			\hline
			\checkmark & & & 22.54 & 0.7273 & 0.1906  \\
			\checkmark & \checkmark & & \textcolor{blue}{22.61} & 0.7342 & 0.1842  \\
			\checkmark & & \checkmark & 22.56 & \textcolor{blue}{0.7408} & \textcolor{red}{0.1470}  \\
			\checkmark & \checkmark & \checkmark & \textcolor{red}{22.73} & \textcolor{red}{0.7420} & \textcolor{blue}{0.1580}  \\
			\bottomrule[1pt]
		\end{tabular}
	\end{center}
	\caption{Ablation study for loss function.}
	\label{tab_ablation_loss}
\end{table}

\begin{figure}[!t]
	\centering
	\begin{subfigure}[b]{0.32\linewidth}
		\centering
		\includegraphics[width=\textwidth]{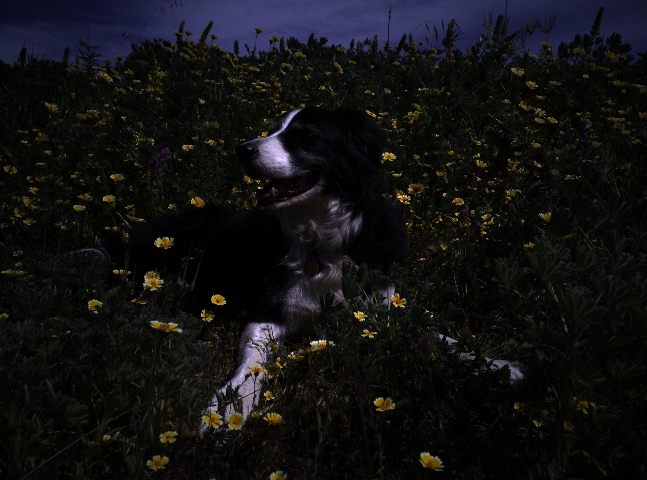}
		\caption{input}
	\end{subfigure}
	\hfill
	\begin{subfigure}[b]{0.32\linewidth}
		\centering
		\includegraphics[width=\textwidth]{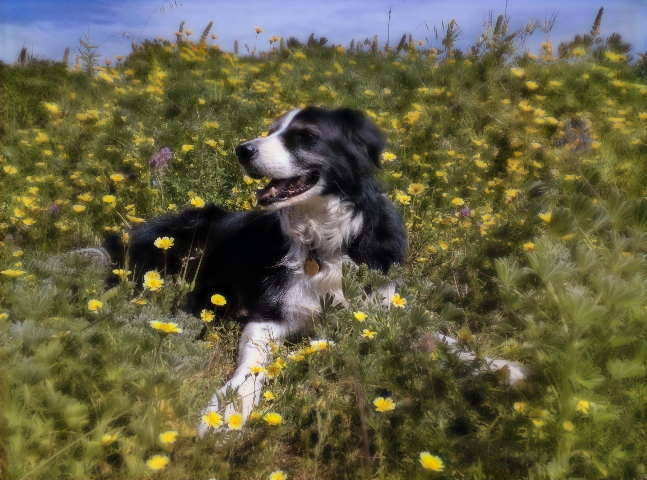}
		\caption{$L_b$}
	\end{subfigure}
	\hfill
	\begin{subfigure}[b]{0.32\linewidth}
		\centering
		\includegraphics[width=\textwidth]{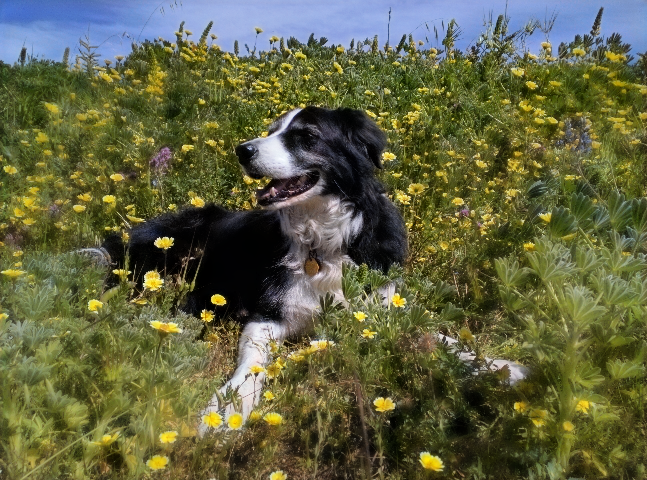}
		\caption{$L_b+L_c$}
	\end{subfigure}
	\begin{subfigure}[b]{0.32\linewidth}
		\centering
		\includegraphics[width=\textwidth]{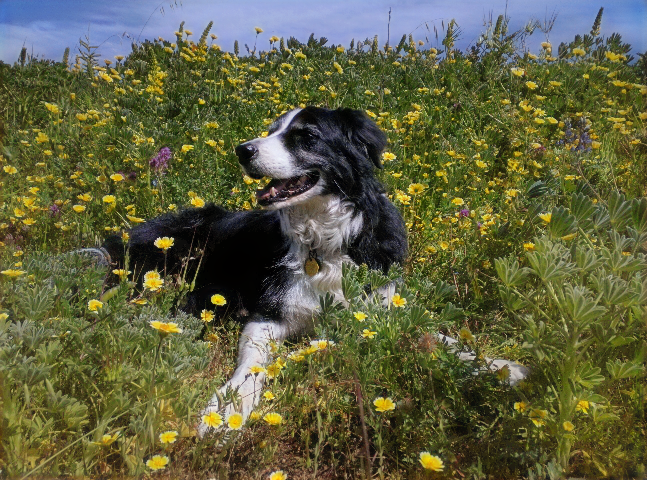}
		\caption{$L_b+L_p$}
	\end{subfigure}
	\hfill
	\begin{subfigure}[b]{0.32\linewidth}
		\centering
		\includegraphics[width=\textwidth]{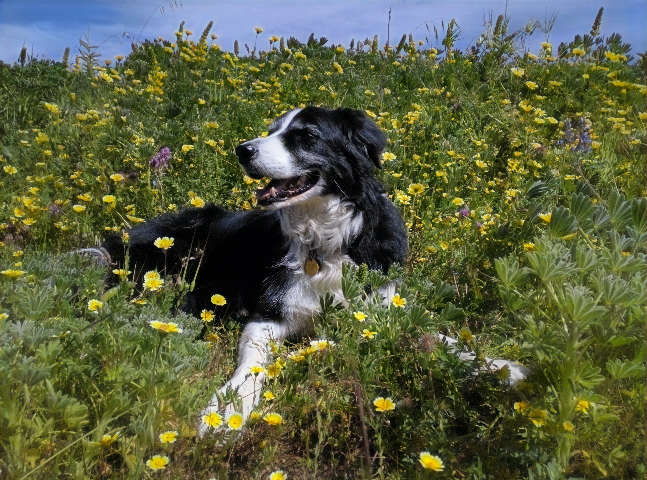}
		\caption{$L_b+L_c+L_p$}
	\end{subfigure}
	\hfill
	\begin{subfigure}[b]{0.32\linewidth}
		\centering
		\includegraphics[width=\textwidth]{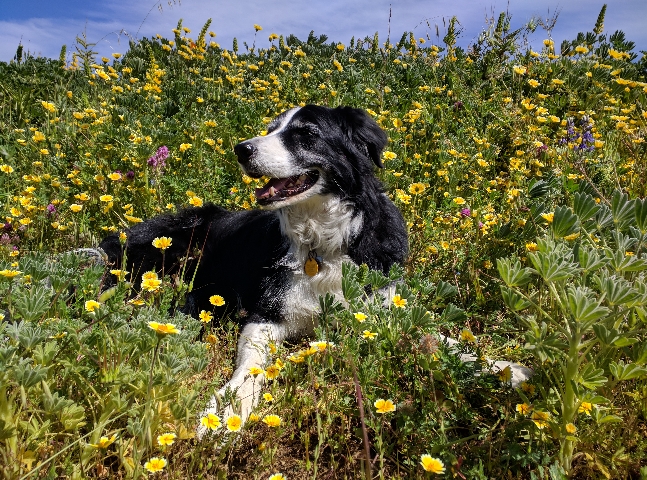}
		\caption{Ground-truth}
	\end{subfigure}
	\caption{Visual results of ablation study on loss functions. (b) is blurry, which means only $L_b$ cannot guarantee satisfied results. (c) looks more vivid and is much closer to ground-turth in color under the supervision of $L_c$, but plants still look fuzzy. By introducing $L_p$, (d) is clearer and sharper in detail like dog hair and grass. With both $L_c$ and $L_p$, (e) is improved significantly in color, detail and local contrast and has the most pleasant perception.}
	\label{fig_ablation_loss}
\end{figure}

\begin{figure}[tb]
	\centering
	\begin{subfigure}[b]{0.23\linewidth}
		\centering
		\includegraphics[width=\textwidth]{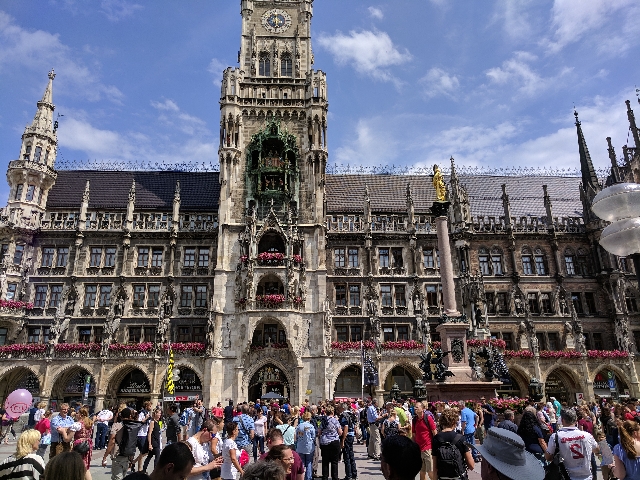}
	\end{subfigure}
	\hfill
	\begin{subfigure}[b]{0.23\linewidth}
		\centering
		\includegraphics[width=\textwidth]{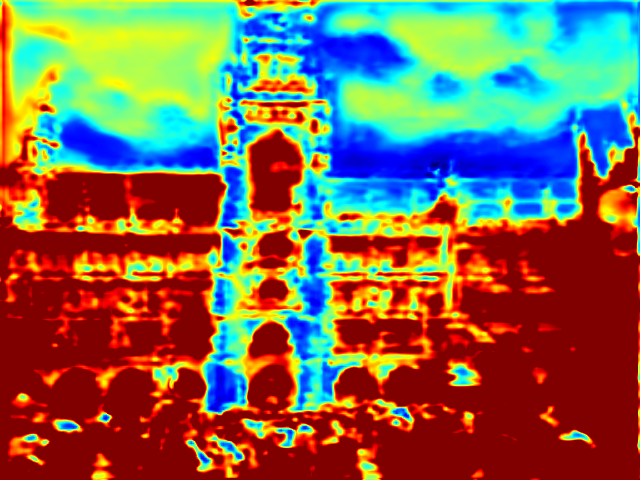}
	\end{subfigure}
	\hfill
	\begin{subfigure}[b]{0.23\linewidth}
		\centering
		\includegraphics[width=\textwidth]{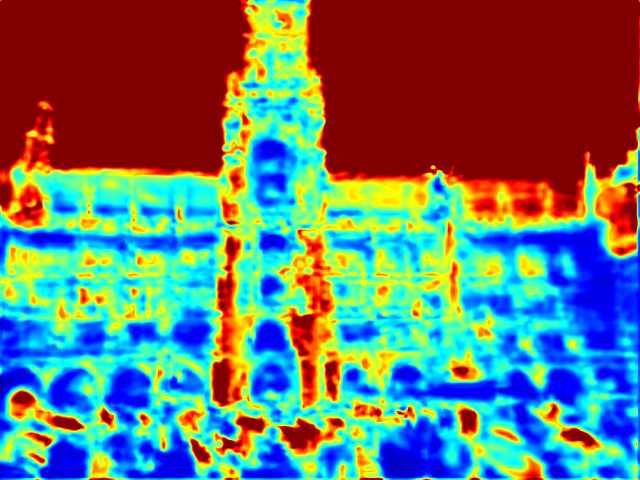}
	\end{subfigure}
	\hfill
	\begin{subfigure}[b]{0.23\linewidth}
		\centering
		\includegraphics[width=\textwidth]{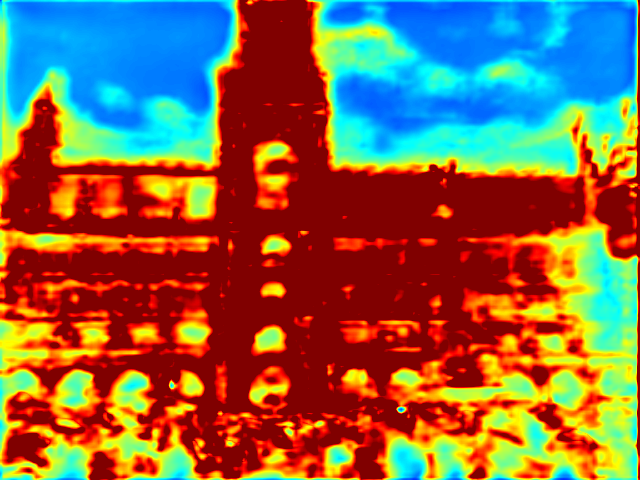}
	\end{subfigure}
	
	\smallskip
	\begin{subfigure}[b]{0.23\linewidth}
		\centering
		\includegraphics[width=\textwidth]{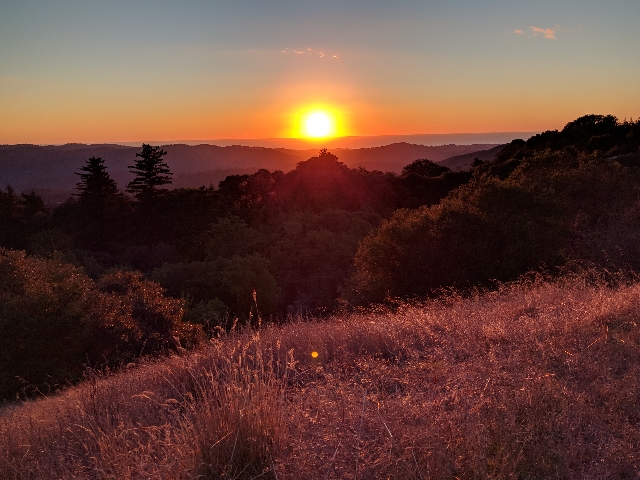}
	\end{subfigure}
	\hfill
	\begin{subfigure}[b]{0.23\linewidth}
		\centering
		\includegraphics[width=\textwidth]{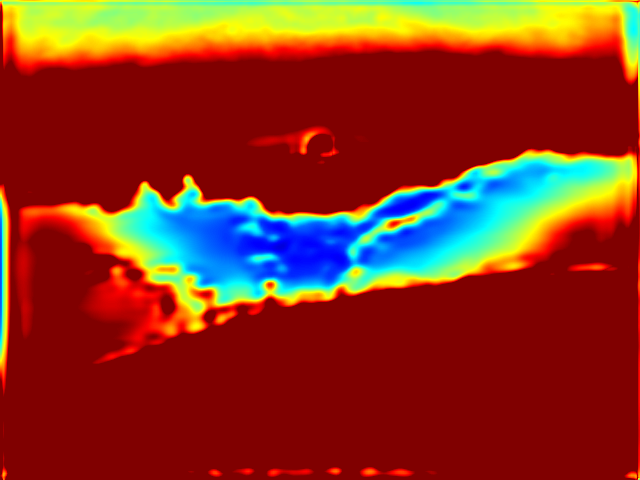}
	\end{subfigure}
	\hfill
	\begin{subfigure}[b]{0.23\linewidth}
		\centering
		\includegraphics[width=\textwidth]{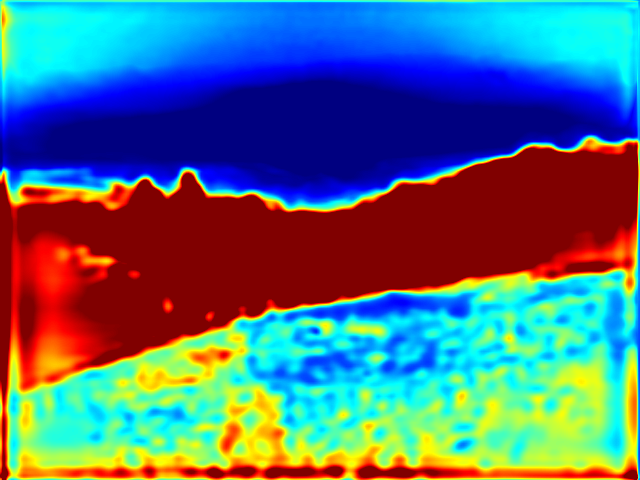}
	\end{subfigure}
	\hfill
	\begin{subfigure}[b]{0.23\linewidth}
		\centering
		\includegraphics[width=\textwidth]{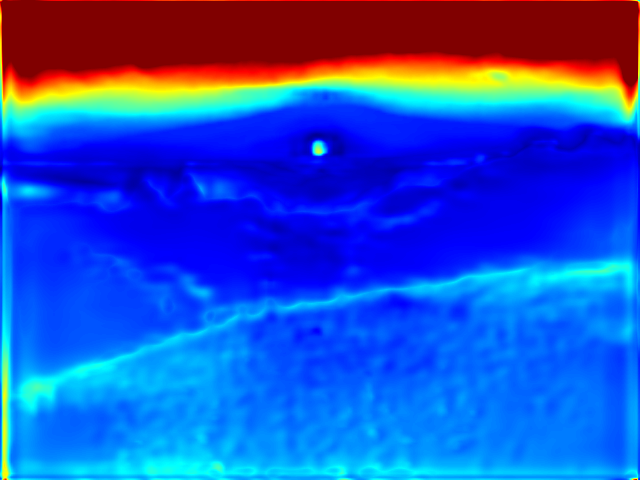}
	\end{subfigure}
	\caption{Visual result of the pixel-aware category weight map. In each row, the first is the ground-truth, and the other three are visualization for different channels. Red pixels for more activated and blue for less activated.}
	\label{fig_cap_map}
\end{figure}

\textbf{Loss function.} The loss function in~\cite{zeng2020learning} is defined as our basic loss ($L_b$), which is a combination of MSE loss, smooth loss and monotonicity loss. We train our models using different combinations of losses to evaluate the influence of our loss function. 

Quantitative results are demonstrated in Table~\ref{tab_ablation_loss}, indicating that the model trained with only $L_b$ gets relatively poor results, and that after the introduction of color loss and perception loss, all 3 metrics are dramatically improved. More analysis can be found in Figure~\ref{fig_ablation_loss}.
\subsection{Analysis of pixel-wise category map}
Some pixel-wise category maps are visualized for analysis. We do not apply any loss on the category map, willing that it can be image adaptive for the network, but not perception adaptive from person’s point of view. On one hand, a perception adaptive category map does not guarantee better performances. In fact, we first apply explicit loss on the category map, but found  1.52dB degradation in PSNR. On the other hand, images can be categorized more flexibly, according to semantic, illumination, or other properties learned by the network itself. For the first row in Figure~\ref{fig_cap_map}, the image may be classified semantically, with three maps indicating person, sky, and building respectively. For the second row, brightness is learned by our weight predictor, where three maps represent middle, low and high illumination areas.

\subsection{Comparison with State-of-the-Arts}

We compare our approach with several SOTA supervised image enhancement methods including DPED~\cite{ignatov2017dslr}, RSGUNet~\cite{huang2018range}, HPEU~\cite{huang2019hybrid}, HDRNet~\cite{gharbi2017deep}, UPE~\cite{wang2019underexposed}, DeepLPF~\cite{moran2020deeplpf}, 3DLUT~\cite{zeng2020learning} on MIT-Adobe FiveK and HDR+ dataset. 

Among these methods, DPED, RSGUNet HPEU and DeepLPF are pixel-level enhancement methods based on ResNet and Unet backbone, while HDRNet and UPE belongs to patch-level methods, and 3DLUT is the image-level method. All of these methods are trained by publicly available source codes with recommended configurations.

\begin{table*}[!t]
	\centering
	\begin{tabularx}{\textwidth}{p{0.12\textwidth} X X X X X X X }
		\toprule[1pt]
		\multirow{2}{100em}{Method} & \multicolumn{3}{c}{480p (~\cite{zeng2020learning})} & \multicolumn{3}{c}{Full resolution (Ours)} \\
		\cmidrule(lr){2-4} \cmidrule(lr){5-7} 
		& PSNR $\uparrow$ & SSIM $\uparrow$ &LPIPS $\downarrow$ &  PSNR $\uparrow$ & SSIM $\uparrow$ &LPIPS $\downarrow$ \\ 
		\hline 
		RSGUNet~\cite{huang2018range} & 22.16 & 0.8382 & 0.0701 &  21.37 & {0.7998} & {0.1861} \\
		DPED~\cite{ignatov2017dslr} & 24.06 & 0.8557 & 0.0935 &  N.A & N.A & N.A \\
		HPEU~\cite{huang2019hybrid} & 24.14 & 0.8754 & 0.0796 &  \textcolor{blue} {22.84} & {0.8356} & 0.2070 \\
		HDRNet~\cite{gharbi2017deep} & 24.22 & 0.8821 & 0.0609 & {22.15} & \textcolor{blue} {0.8403} & \textcolor{blue}{0.1823} \\
		UPE~\cite{wang2019underexposed} & 21.35 & 0.8191 & 0.1162 &  20.03 & 0.7841 & 0.2523 \\
		DeepLPF~\cite{moran2020deeplpf} & \textcolor{blue}{25.29} & \textcolor{red}{0.8985} & \textcolor{blue}{0.0528} &  N.A & N.A & N.A \\
		3DLUT~\cite{zeng2020learning} & 25.24 & 0.8864 & 0.0530 & {22.27} & {0.8368} & {0.1832} \\
		Ours & \textcolor{red}{25.50} & \textcolor{blue}{0.8904} & \textcolor{red}{0.0512} &  \textcolor{red}{23.17} & \textcolor{red}{0.8636} & \textcolor{red}{0.1451} \\
		\bottomrule[1pt]
	\end{tabularx}
	\caption{Quantitative results on MIT-Adobe FiveK. N.A. means the result is not available due to insufficient memory of GPU.}
	\label{tab_MIT_comp_sota}
\end{table*}

\begin{table*}[!t]
	\centering
	\begin{tabularx}{\textwidth}{p{0.12\textwidth} X X X X X X X X X}
		\toprule[1pt]
		\multirow{2}{100em}{Method} & \multicolumn{3}{c}{480p (~\cite{zeng2020learning})} & \multicolumn{3}{c}{480p (Ours)} & \multicolumn{3}{c}{Full resolution (Ours)} \\
		\cmidrule(lr){2-4} \cmidrule(lr){5-7} \cmidrule(lr){8-10}
		& PSNR $\uparrow$ & SSIM $\uparrow$ &LPIPS $\downarrow$ & PSNR $\uparrow$ & SSIM $\uparrow$ &LPIPS $\downarrow$ & PSNR $\uparrow$ & SSIM $\uparrow$ &LPIPS $\downarrow$ \\ 
		\hline 
		RSGUNet~\cite{huang2018range} & 25.03 & 0.8903 & 0.0751 & 19.47 & 0.6725 & 0.2023 & 19.76 & \textcolor{blue}{0.6945} & \textcolor{red}{0.2474} \\
		DPED~\cite{huang2019hybrid} & 25.61 & 0.9098 & 0.0806 & 21.04 & 0.6834 & 0.2389 & \textcolor{blue}{20.97} & {0.6798} & 0.3264 \\
		HPEU~\cite{ignatov2017dslr} & 25.12 & 0.8733 & 0.1193 & 21.08 & 0.7168 & 0.2198 & 19.43 & 0.6679 & 0.3923 \\	
		HDRNet~\cite{gharbi2017deep} & 26.72 & 0.9024 & 0.0758 & 20.95 & 0.6914 & 0.2310 & {20.04} & 0.6378 & 0.3559 \\
		UPE~\cite{wang2019underexposed} & 24.96 & 0.8655 & 0.1144 & 19.87 & 0.6445 & 0.2693 & 19.42 & 0.5516 & 0.4568 \\
		DeepLPF~\cite{moran2020deeplpf} & \textcolor{blue}{27.44} & \textcolor{red}{0.9388} & \textcolor{blue}{0.0496} & \textcolor{blue}{22.13} & \textcolor{red}{0.7467} & \textcolor{blue}{0.1986} & N.A & N.A & N.A \\
		3DLUT~\cite{zeng2020learning} & 23.59 & 0.8844 & 0.1057 & 19.91 & 0.6567 & 0.2455 & 19.88 & 0.5942 & 0.4089 \\
		Ours & \textcolor{red}{28.29} & \textcolor{blue}{0.9279} & \textcolor{red}{0.0562} & \textcolor{red}{22.73} & \textcolor{blue}{0.7420} & \textcolor{red}{0.1580} & \textcolor{red}{22.56} & \textcolor{red}{0.6996} & \textcolor{blue}{0.2808} \\
		\bottomrule[1pt]
	\end{tabularx}
	\caption{Quantitative results on HDR+ dataset. N.A. means the result is not available due to insufficient memory of GPU.}
	\label{tab_HDR+_comp_sota}
\end{table*}

\begin{figure*}[!htb]
	\centering
	\begin{subfigure}[b]{0.24\linewidth}
		\centering
		\includegraphics[width=\textwidth]{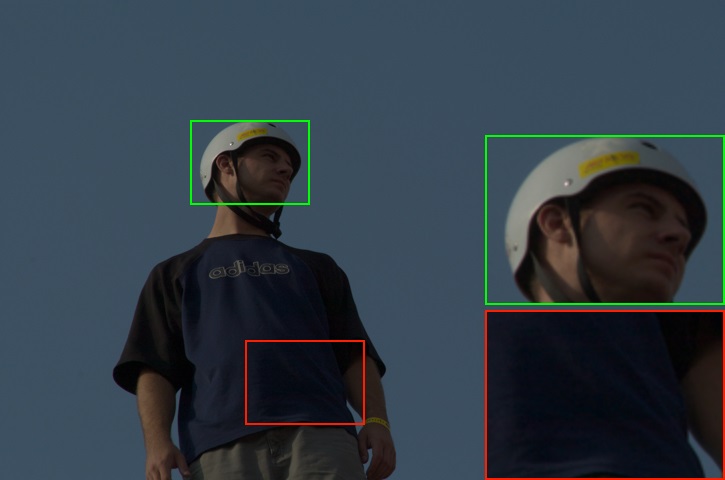}
		\caption{Input}
	\end{subfigure}
	\hfill
	\begin{subfigure}[b]{0.24\linewidth}
		\centering
		\includegraphics[width=\textwidth]{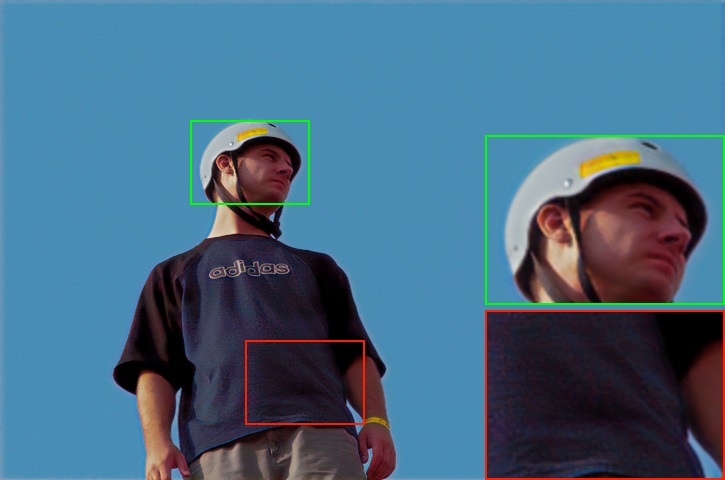}
		\caption{DPED~\cite{ignatov2017dslr}}
	\end{subfigure}
	\hfill
	\begin{subfigure}[b]{0.24\linewidth}
		\centering
		\includegraphics[width=\textwidth]{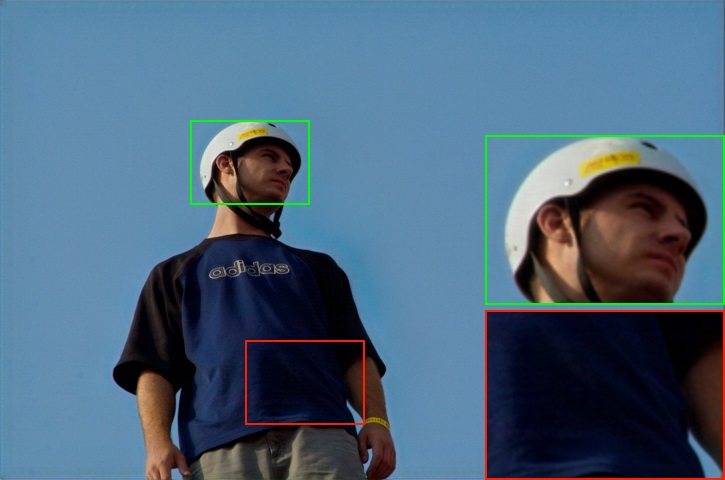}
		\caption{HPEU~\cite{huang2019hybrid}}
	\end{subfigure}
	\hfill
	\begin{subfigure}[b]{0.24\linewidth}
		\centering
		\includegraphics[width=\textwidth]{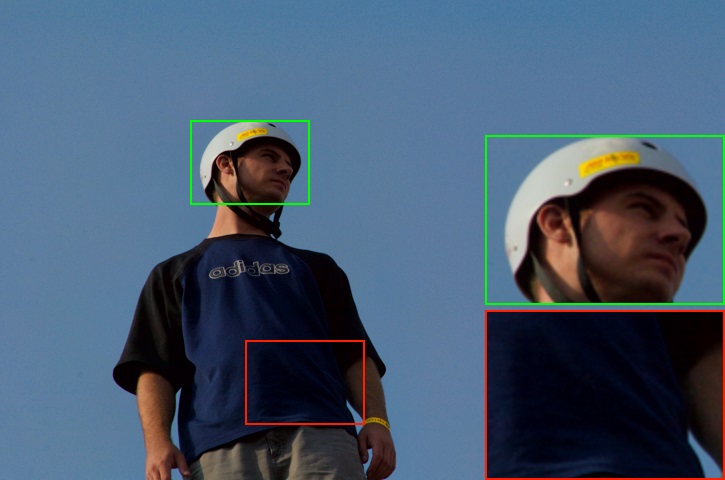}
		\caption{HDRNet~\cite{gharbi2017deep}}
	\end{subfigure}
	\vfill
	\begin{subfigure}[b]{0.24\linewidth}
		\centering
		\includegraphics[width=\textwidth]{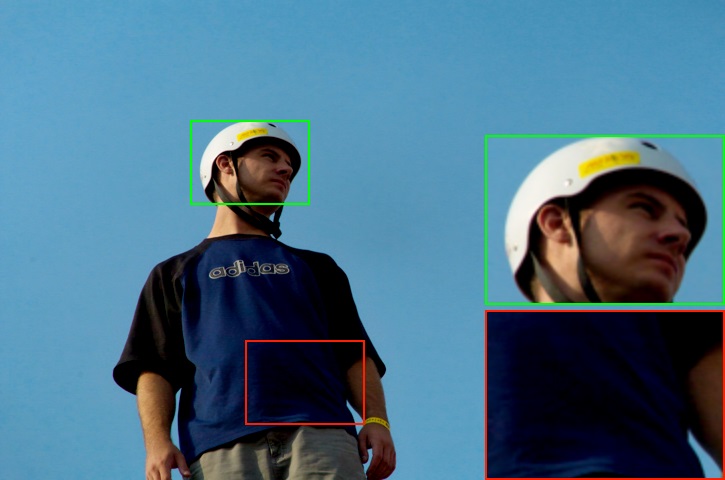}
		\caption{DeepLPF~\cite{moran2020deeplpf}}
	\end{subfigure}
	\hfill
	\begin{subfigure}[b]{0.24\linewidth}
		\centering
		\includegraphics[width=\textwidth]{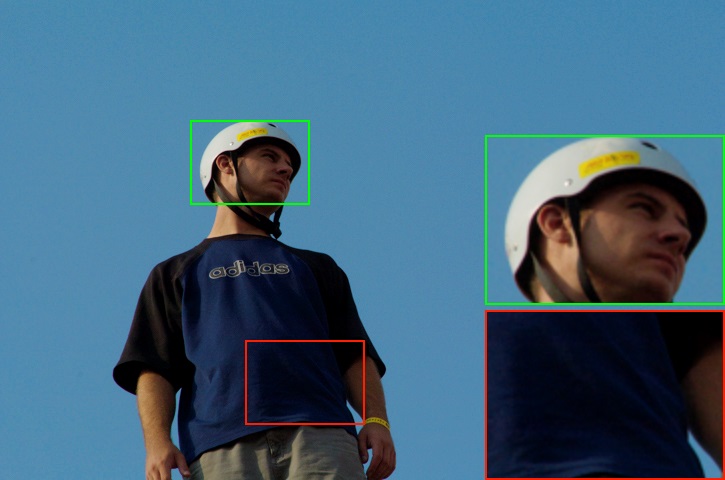}
		\caption{3DLUT~\cite{zeng2020learning}}
	\end{subfigure}
	\hfill
	\begin{subfigure}[b]{0.24\linewidth}
		\centering
		\includegraphics[width=\textwidth]{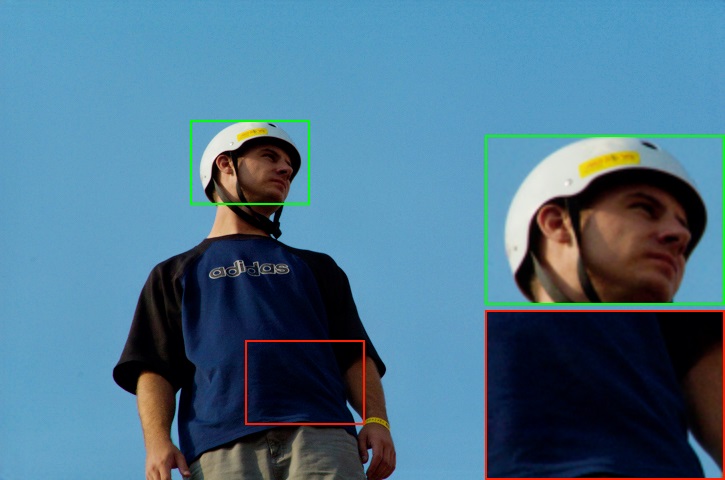}
		\caption{Ours}
	\end{subfigure}
	\hfill
	\begin{subfigure}[b]{0.24\linewidth}
		\centering
		\includegraphics[width=\textwidth]{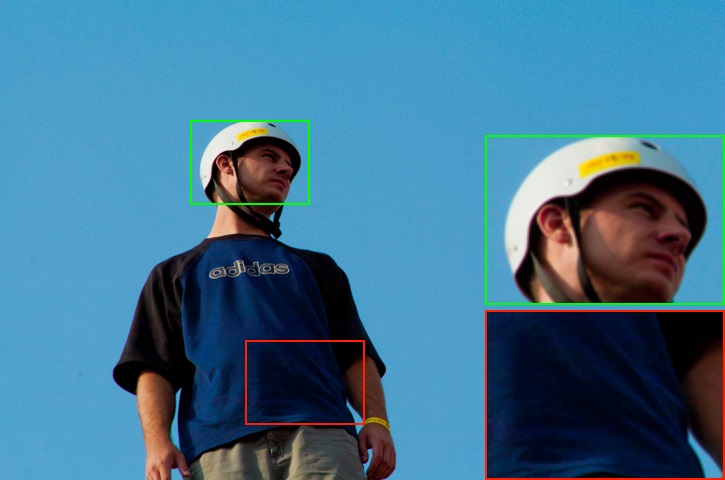}
		\caption{Ground-truth}
	\end{subfigure}
	\caption{Results comparison on 'a3909' of 480p MIT-Adobe FiveK dataset. Our result outperforms all other methods both in color and detail. For example, (b) is bit of red on face and cloths, (c), (d) and (f) suffer from insufficient saturation in background areas, (c) has obvious contour artifacts around the person, and (e) is a bit darker compared with our result and some textures on cloths are lost.}
	\label{fig_sota_mit5k}
\end{figure*}

\begin{figure*}[!htb]
	\centering
	\begin{subfigure}[b]{0.24\linewidth}
		\centering
		\includegraphics[width=\textwidth]{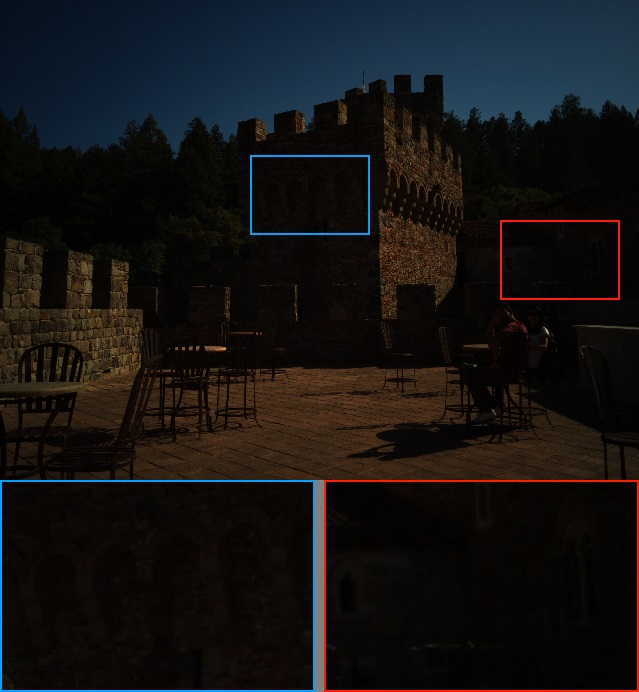}
		\caption{Input}
	\end{subfigure}
	\hfill
	\begin{subfigure}[b]{0.24\linewidth}
		\centering
		\includegraphics[width=\textwidth]{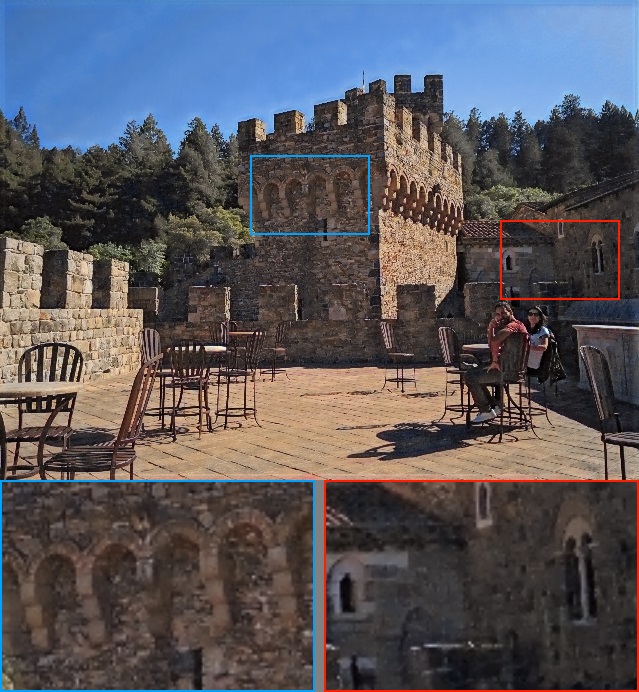}
		\caption{DPED~\cite{ignatov2017dslr}}
	\end{subfigure}
	\hfill
	\begin{subfigure}[b]{0.24\linewidth}
		\centering
		\includegraphics[width=\textwidth]{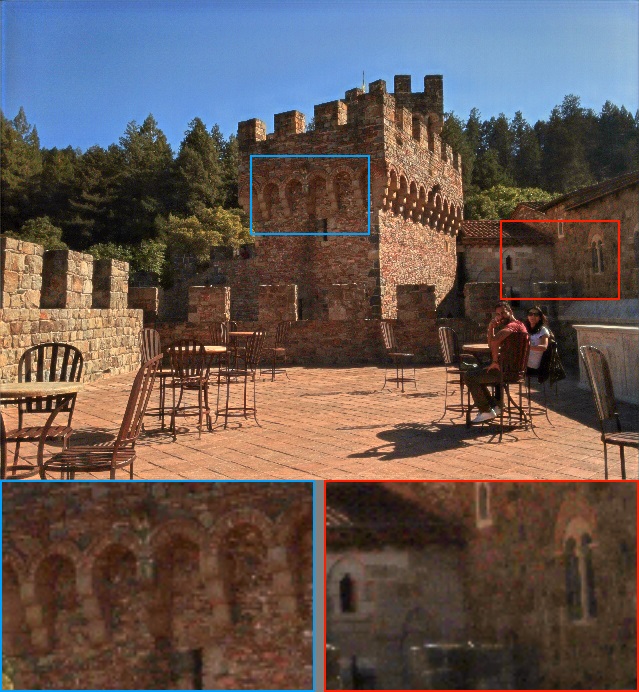}
		\caption{HPEU~\cite{huang2019hybrid}}
	\end{subfigure}
	\hfill
	\begin{subfigure}[b]{0.24\linewidth}
		\centering
		\includegraphics[width=\textwidth]{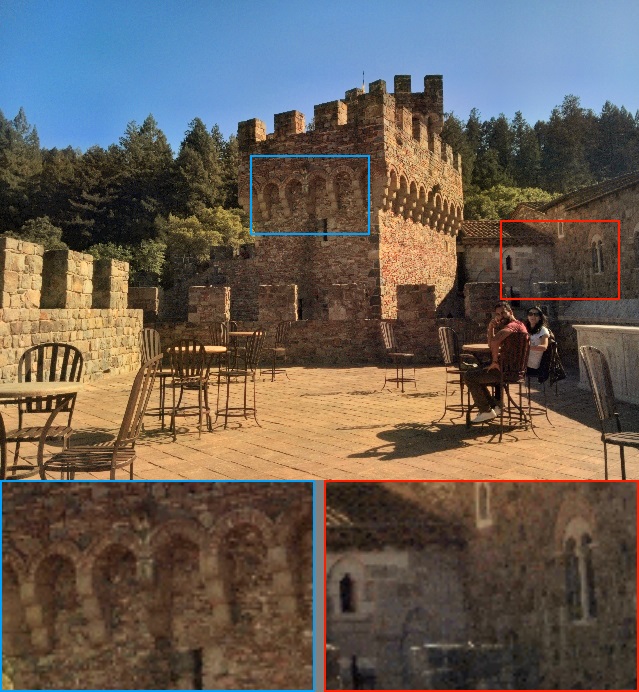}
		\caption{HDRNet~\cite{gharbi2017deep}}
	\end{subfigure}
	\vfill
	\begin{subfigure}[b]{0.24\linewidth}
		\centering
		\includegraphics[width=\textwidth]{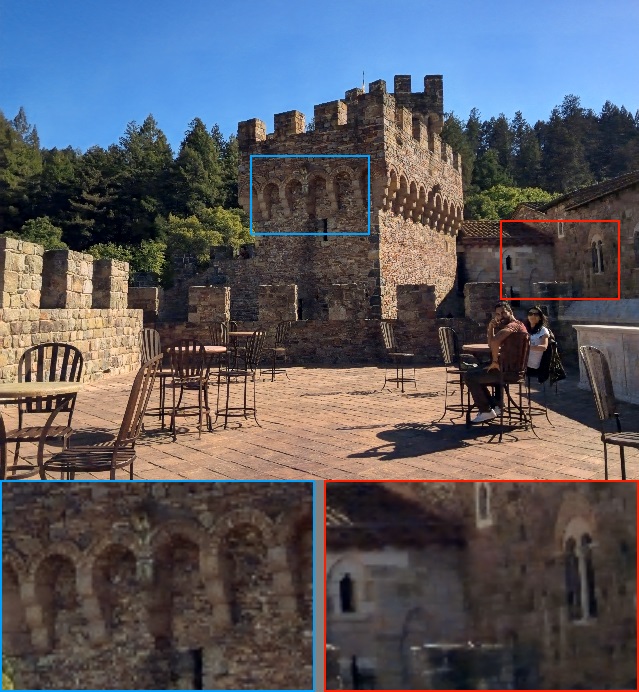}
		\caption{DeepLPF~\cite{moran2020deeplpf}}
	\end{subfigure}
	\hfill
	\begin{subfigure}[b]{0.24\linewidth}
		\centering
		\includegraphics[width=\textwidth]{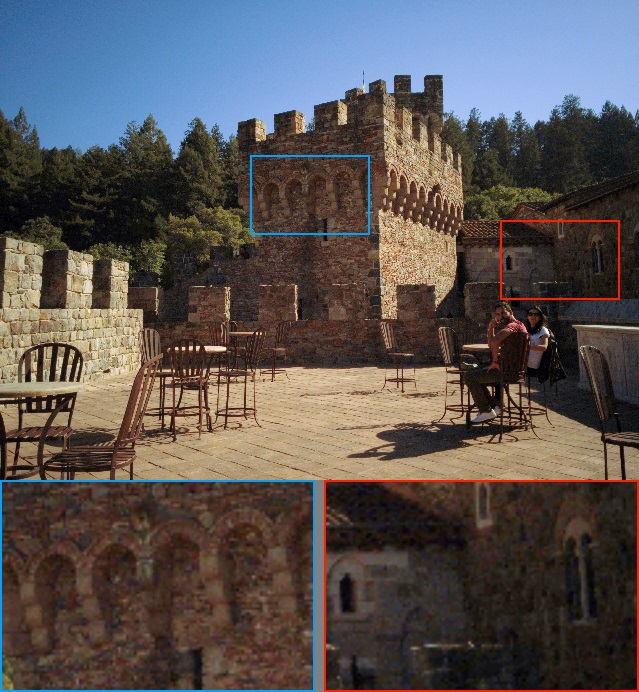}
		\caption{3DLUT~\cite{zeng2020learning}}
	\end{subfigure}
	\hfill
	\begin{subfigure}[b]{0.24\linewidth}
		\centering
		\includegraphics[width=\textwidth]{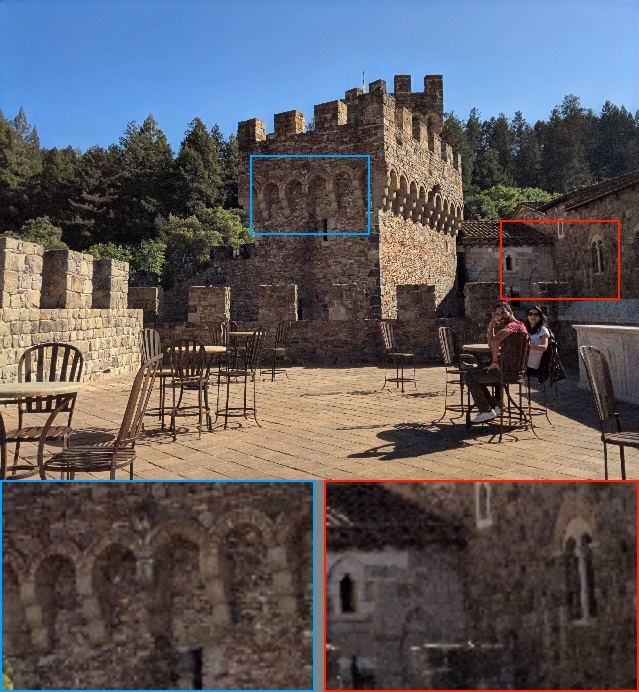}
		\caption{Ours}
	\end{subfigure}
	\hfill
	\begin{subfigure}[b]{0.24\linewidth}
		\centering
		\includegraphics[width=\textwidth]{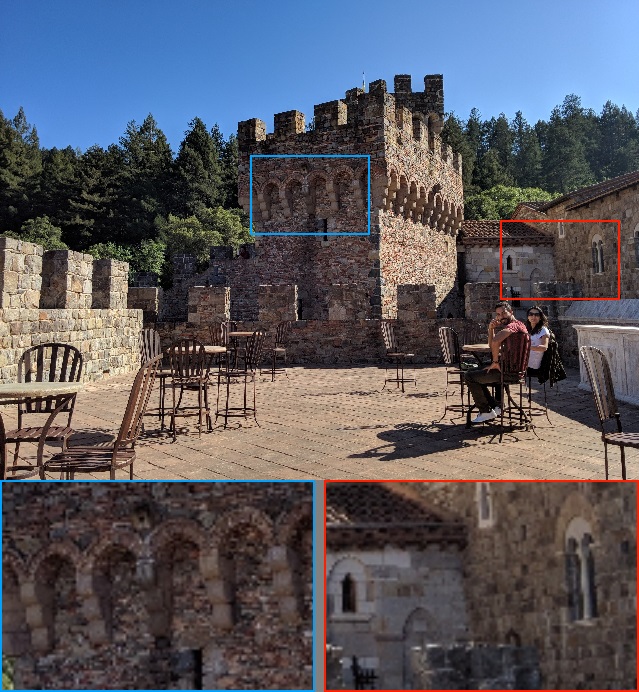}
		\caption{Ground-truth}
	\end{subfigure}
	\caption{Results comparison on '5a9e\_20150403\_162152\_482' of our 480p HDR+ burst photography dataset. Results from (c), and (d) are both slightly yellow especially in the blue block area. (b) shows severe bending artifact in sky. The result of (e) is blurred in the red block area. Our model is much closer to the ground-truth, with much better color, clearer texture and less artifacts. In addition, owning to the pixel-aware category information, our model is able to enhance local areas differently, while traditional (f) can only enhance the whole image uniformly with local area in red block remaining dark. }
	\label{fig_sota_hdrplus}
\end{figure*}

As shown in Table~\ref{tab_MIT_comp_sota}, our approach outperforms other methods in terms of PSNR and LPIPS on MIT-Adobe FiveK. For SSIM on 480p, our result is a bit lower ($<$1\%) than DeepLPF, but all other metrics are much better than DeepLPF. Particularly, due to the large memory consumption, the complicated DeepLPF algorithm cannot be applied on full resolution image(i.e., 4K resolution image). Similar result can be seen in Table~\ref{tab_HDR+_comp_sota} on HDR+ dataset. Our model outperforms the second best model by 0.85dB and 1.59dB on 480p and full resolution respectively. The performance gap between our 480p HDR+ dataset and ~\cite{zeng2020learning} may be that the number of our HDR+ dataset is much larger than ~\cite{zeng2020learning}, resulting more serious disalignment. Mapping for HDR+ dataset is more locally complicated that it contains scenarios with wider dynamic range and more various illumination. Hence, our spatial-aware 3D LUTs with pixel-wise category map is more adaptive to those local variant transformations and have an evident improvement. On all datasets, our method achieves great improvement compared with basic 3DLUT method in all criterions. As the visual result shown in Figure~\ref{fig_sota_mit5k} and Figure~\ref{fig_sota_hdrplus}, it indicates that our results are more visually pleasant, and are closer to the ground-truth. More visual results can be found in supplementary material.

Apart from pleasant visual perception, our method is efficient for both low and high resolution images. Table~\ref{tab_running_time} shows the inference time for different models with input size $640 \times 480$, $1920 \times 1080$, and $3840 \times 2160$ on 32GB NVIDIA V100 GPU. Our model takes a bit longer running time when compared with 3DLUT, but it is about two-order faster than all other methods. Additionally, it only takes about 4 ms for our model to process a 4K resolution image, which exceeds the requirement of real-time processing by a large amount. The high efficiency mainly owns to the characteristic that our CNN network generates two kinds of weight information from a low resolution input, and that the spatial-aware interpolation sensitive to the image size is greatly accelerated via customized CUDA codes. Thus, running time for our spatial-aware 3D LUTs remains approximately unchanged, while other competing methods except 3DLUT takes exponentially longer time as the resolution gets higher.

\begin{table}[tb]
	\begin{center}
		\begin{tabular}{l c c c}
			\toprule[1pt]
			Resolution & 640x480 & 1920x1080 & 3840x2160 \\
			\hline
			RSGUNet~\cite{huang2018range} & 6.12 & 37.16 & 158.4 \\
			DPED~\cite{ignatov2017dslr} & 58.63 & 408.5 & 1702\\
			HPEU~\cite{huang2019hybrid} & 5.75 & 36.88 & 189.1\\
			HDRNet~\cite{gharbi2017deep} & 3.82 & 31.68 & 142.2\\
			UPE~\cite{wang2019underexposed} & 4.16 & 33.3 & 133.26\\
			DeepLPF~\cite{moran2020deeplpf} & 40.38 & 146.8 & N.A.\\
			3DLUT~\cite{zeng2020learning} & \textcolor{red}{1.13} & \textcolor{red}{1.19} & \textcolor{red}{1.22}\\
			Ours & \textcolor{blue}{2.27} & \textcolor{blue}{2.34} & \textcolor{blue}{4.39}\\
			\bottomrule[1pt]
		\end{tabular}
	\end{center}
	\caption{Running time (in millisecond) comparison between our approach and current SOTA CNN-based methods on different resolutions. All methods are tested on NVIDIA V100 GPU. N.A. means the result is not available due to insufficient memory of GPU. }
	\label{tab_running_time}
\end{table}

\section{Discussion and Conclusion}
Traditional 3DLUTs interpolate merely through RGB colors, leading to pool local contrast, while bilateral grids interpolate through luminance and space. However, this leads to more computational overheads and longer inference time for bilateral grids, since they are strongly coupled with slicing operation and a guide map of input’s resolution. Computation of this guide map is heavy and time consuming, especially for high resolution inputs. Table 4 shows that bilateral grids in HDRNet are sensitive to resolution, with 0.91dB difference in PSNR for inferring 480p and 4K images.We can conclude that for a fixed grid size, performance for HDRNet decreases as input’s resolution gets larger.

Our proposed spatial-aware 3D LUTs, on the other hand, produce charming results with good local contrast in high efficiency. Its key idea is constructing spatial-aware 3D LUTs with pixel-wise category map to improve the robustness in local regions for traditional 3D LUT. Further, we design a two-head weight predictor that generates different level of category information, enabling our network to be image-level scenario and pixel-wise category adaptive. Extensive experiments on public datasets demonstrate the superiority of our method against many SOTA methods on both performance and efficiency.

\end{document}